\documentclass{article}

\usepackage{microtype}
\usepackage{graphicx}
\usepackage{subfigure}
\usepackage{booktabs} 

\usepackage{hyperref}

\usepackage[accepted]{icml2023}

\usepackage{amsmath}
\usepackage{amssymb}
\usepackage{mathtools}
\usepackage{amsthm}

\usepackage[capitalize,noabbrev]{cleveref}

\theoremstyle{plain}

\theoremstyle{definition}

\theoremstyle{remark}

\usepackage[textsize=tiny]{todonotes}

\icmltitlerunning{Mitigating Label Bias via Decoupled Confident Learning}

\begin{document}

\twocolumn[
\icmltitle{Mitigating Label Bias via Decoupled Confident Learning}



\icmlsetsymbol{equal}{*}

\begin{icmlauthorlist}
\icmlauthor{Yunyi Li}{yyy}
\icmlauthor{Maria De-Arteaga}{yyy}
\icmlauthor{Maytal Saar-Tsechansky}{yyy}
\end{icmlauthorlist}

\icmlaffiliation{yyy}{University of Texas at Austin, Austin, TX, USA}

\icmlcorrespondingauthor{Yunyi Li}{Yunyi.Li@mccombs.utexas.edu}

\icmlkeywords{Machine Learning, ICML}

\vskip 0.3in
]



\printAffiliationsAndNotice{}  

\begin{abstract}
Growing concerns regarding algorithmic fairness have led to a surge in methodologies to mitigate algorithmic bias. However, such methodologies largely assume that observed labels in training data are correct. This is problematic because bias in labels is pervasive across important domains, including healthcare, hiring, and content moderation. In particular, human-generated labels are prone to encoding societal biases. While the presence of labeling bias has been discussed conceptually, there is a lack of methodologies to address this problem. We propose a pruning method---\textbf{De}coupled \textbf{Co}nfident \textbf{Le}arning (DeCoLe)---specifically designed to mitigate label bias. After illustrating its performance on a synthetic dataset, we apply DeCoLe in the context of hate speech detection, where label bias has been recognized as an important challenge, and show that it successfully identifies biased labels and outperforms competing approaches.
\end{abstract}

\section{Introduction}\label{Intro}
The rapid advancement of AI technologies has sparked important discussions on fairness, equity, and ethics, as AI becomes more integrated into our daily lives. In particular, there has been a surge in the development and application of methodologies to mitigate algorithmic bias. However, most of the methods developed to address algorithmic bias often rely on a critical assumption: the observed labels used to train these systems are accurate. This assumption, while convenient, is often violated, given that bias is often embedded in the training labels, which can result in undesirable consequences that further exacerbate bias \cite{li2022more}. 
 
\emph{Label bias} refers to a systematic disparity between the ground truth labels intended to train an AI system and the observed labels, such that the relationship underlying the mismatch differs across groups~\cite{li2022more}. This type of bias is particularly prominent when labels are generated by humans. For example, African American English (AAE) Twitter posts are disproportionately labeled as toxic by crowd-sourcing annotators, despite those posts being understood as non-toxic by AAE speakers~\cite{sap2019risk}. In general, such systematic disparity between ground truth and observed labels can result from a wide range of causes, including historical discrimination, human cognitive bias, or social inequalities. For these reasons, label bias has been a growing concern across domains, including but not limited to healthcare~\cite{obermeyer2019dissecting, philbin1998influence}, criminal justice~\cite{fogliato2021validity}, and hiring~\cite{hunter1979differential}.


Exiting methods for mitigating algorithmic bias often operate under the assumption that the observed labels are accurate and reliable for training and evaluation. Particularly, this is true for measures of bias that are derived from the confusion matrix, such as equalized odds~\cite{hardt2016equality}. However, this assumption can be highly problematic since machine learning models can inadvertently perpetuate existing biases encoded in the biased labels. Procedures for mitigating bias and evaluating model performance with respect to biased observed labels can result in misleadingly positive performance on testing data, thereby exacerbating the existing bias without being aware of it \cite{li2022more}. 

The presence of label bias in many datasets across different important domains introduces a new challenge: How can we identify instances in the data that are highly likely to be mislabeled, such that removing those instances can decrease label bias in the data?

A large body of work dealing with noisy labels addresses different challenges related to uncertainty in labels. Such works typically make assumptions about the noise form, such as uniform or class-conditional noise~\cite{angluin1988learning}. The common assumption of class-conditional noise is that label noise depends only on the latent true class. This assumption is commonly used~\cite{goldberger2017training, sukhbaatar2014training}, and may be reasonable under some circumstances. However, the noise structure can be closely related to group membership, which can lead to label bias. For instance, in the context of toxicity detection, false positive labels may be more common among posts written in AAE. Moreover, instances from different groups can exhibit distinct relationships between covariates and labels---a phenomenon sometimes referred to as differential subgroup validity---, and under such circumstances, predictive models may be dominated by the relationships that hold true for the majority~\citep{chouldechova2020snapshot}. For example, hateful posts targeting different groups often take very different forms~\cite{gupta2023same}. Under such circumstances, there is a potential risk that models that mitigate label noise through a ``one size fits all" approach may fail to correctly identify erroneous labels affecting minority groups.  

We propose a novel pruning method---Decoupled Confident Learning (DeCoLe)---specifically designed to mitigate label bias. The goal is to identify instances for which the label is likely to be mistaken, so that such instances may be pruned. We estimate the group and class-conditional label uncertainty by training decoupled classifiers~\citep{dwork2018decoupled}, and perform group-specific pruning based on the principles of estimating incorrect labels in classification \cite{northcutt2021confident, elkan2001foundations, forman2005counting}. 
Notably, DeCoLe is a model-agnostic, \textbf{data-centric} algorithm, making it a general-purpose framework that can be applied to identify erroneous labels and generate a dataset with reduced bias.

The remaining sections are structured as follows: we briefly review the extant research on algorithmic fairness and label bias, bias in human-generated labels, and noise mitigation in \cref{lit_review}.  We then formalize the problem of label bias mitigation with the presence of asymmetric label errors in \cref{preliminary}, and propose Decoupled Confident Learning (DeCoLe) to prune biased labels in \cref{DeCoLe}. Next, we illustrate and validate the performance of DeCoLe on a synthetic dataset in~\cref{sec:synthetic}, followed by a real-world evaluation in the context of hate speech label bias mitigation in \cref{Hate_Speech_describ}. We conclude the paper with a discussion of future research directions in \cref{conclusion}.

\medskip

\section{Related literature} \label{lit_review}
\textbf{Algorithmic fairness and label bias}
While algorithmic fairness has garnered substantial attention, most bias mitigation strategies and metrics of fairness are primarily concerned with inductive bias, and assume that labels available for training are reliable~\citep{chouldechova2020snapshot}. For instance, approaches that equalize errors or other metrics derived from the confusion matrix, do so by evaluating performance with respect to the observed label~\cite{mitchell2018prediction}. \emph{Label bias} has increasingly emerged as a concern. However, most work has centered on characterizing or conceptualizing it. \citet{li2022more} provide an overview of different types of label bias within supervised learning systems and empirically demonstrates that collecting more data can exacerbate bias if label bias is overlooked. \citet{fogliato2020fairness} find that even small biases in observed labels can produce disparate performance across races of recidivism risk assessment tools, and \citet{akpinar2021effect} show that differential rates in crime reporting can lead to bias in predictive policing systems. Label bias has also been identified as a potential problem in other contexts such as healthcare~\cite{obermeyer2019dissecting}, child maltreatment hotline screenings~\cite{de2021leveraging}, hiring~\cite{hunter1979differential}, and offensive language detection~\cite{sap2019risk}.

\textbf{Bias in human-generated labels} With the rise of crowd-sourcing services~\cite{howe2008crowdsourcing}, such as Amazon Mechanical Turk, 
researchers have noted the risks of annotator cognitive biases ~\cite{eickhoff2018cognitive, draws2021checklist} and stereotyping in annotator judgments ~\cite{otterbacher2015crowdsourcing}.
Expert-generated labels can also reflect biases. For example, in healthcare, the quality of pain assessment and treatment recommendations can be undermined by provider biases ~\cite{hoffman2016racial}. For a comprehensive review of label bias and bias in human-generated labels, please refer to \cite{li2022more}.

\textbf{Noise mitigation}
A large stream of work on dealing with noisy labels has proposed aggregating multiple noisy labelers' opinions to reduce the noise in labels~\citep{zhang2016learning}, as well as learning probabilistic models to jointly estimate labelers' quality and gold standard labels~\cite{snow2008cheap, smyth1994inferring, dawid1979maximum, whitehill2009whose, welinder2010multidimensional, yan2010modeling}. Another branch of work focuses on \textit{learning from noisy labels} that do not require labelers' information. Such approaches have investigated training models on noisy datasets through loss reweighting ~\cite{shu2019meta}, surrogate loss~\cite{natarajan2013learning}, co-teaching~\cite{han2018co} and normalized loss functions ~\cite{ma2020normalized}. These approaches tackle the issue of noisy labels by proposing novel model architectures or modifications to the loss function during training. Importantly, methods for learning from noisy labels often assume a particular learning framework and noise structure and cannot be directly adapted to learn any arbitrary model from the data. In particular, they do not consider label bias and assume the noise is either random or solely conditioned on the class. 

DeCoLe builds upon a stream of work that estimates incorrect labels in binary classification~\cite{elkan2001foundations, forman2005counting}, and closely related to the work by \citet{northcutt2021confident}, which focuses on estimating label uncertainty using confident learning (CL)~\citep{northcutt2021confident}. Both CL and DeCoLe are model-agnostic and data-centric, and focus on generating cleaner data. However, it is important to note that the aforementioned works and approaches assume constrained forms of noise such as uniform or class-conditional noise~\cite{angluin1988learning}, excluding the shared societal biases and disregarding fairness considerations during model evaluation. In contrast, DeCoLe specifically addresses the issue of bias in labels, and relaxes the class conditional noise, motivated by the fact that in many cases the noise structure is conditioned on both group and class, as we discussed in \cref{Intro}. To the best of our knowledge, DeCoLe is the pioneering pruning method designed to address the group and class-conditioned label noise. By acknowledging this, DeCoLe aims to mitigate label bias and improve the fairness characteristics of the data post-pruning.

\medskip

\section{Methodology}
In this section, we propose Decoupled Confident Learning (DeCoLe), a pruning approach to mitigate label bias. We first formally introduce the problem of group and class conditioned noise in section \cref{preliminary}. In section \cref{DeCoLe}, we describe key steps in DeCoLe, and present the proposed algorithm.  

\subsection{Preliminaries} \label{preliminary}
In the context of binary classification with possible biased labels, let $\boldsymbol{D} \coloneqq (\boldsymbol{x}, \tilde y) ^n$ denote the dataset of $n$ examples $\boldsymbol{x}$ with associated observed labels $\tilde y \in \{0,1\}$.  We assume there is a group membership indicator $g \in \boldsymbol{x}$, which is typically a pre-defined categorical attribute based on sensitive feature(s). For example, group membership may denote attributes such as age, gender, race, or an intersection of multiple of these. 

Let $\boldsymbol{D} \coloneqq (\boldsymbol{x}, \tilde y) ^n$ be an observed dataset. We suppose there exists a group and class conditional noisy labeling process that results in bias in observed labels $\tilde y$. Let $y^*$ be the latent, unbiased labels (``ground truth labels''). For each group $g_i$, $i \in \{0,...k\} $ where $g_i$ refers to a specific value of $g$, let $\pi_{1\_{g_i}}$ be the fraction of positive instances in group $g_i$ that has been mislabeled as negative, and $\pi_{0\_{g_i}}$ be the fraction of negative instances in group $g_i$ that has been mislabeled as positive. Formally:
\begin{align*}
    &\pi_{1\_{g_i}} = P(\tilde y = 0| y^* = 1, g = g_i)\\
    &\pi_{0\_{g_i}} = P(\tilde y = 1| y^* = 0, g = g_i)
\end{align*}
Label bias occurs when there is a disparity in either $\pi_{1\_{g_i}}$ or $\pi_{0\_{g_i}}$, or in both, accross different groups $g_i$. 
We formulate the problem using a binary classification setting, while allowing multi-categorical group memberships. However, the proposed approach can be extended to a multi-class classification setting. Our goal is to provide a generic approach that can be used to prune erroneous labels in a training dataset, whenever noise is both group and class conditioned. Additionally, we account for the fact that there may be differential subgroup validity, i.e. relationships between the covariates $\boldsymbol{x}$ and the target label $y^*$ may vary across groups. By tackling this, we provide a methodology to mitigate the label bias problem and prevent ML from propagating existing prejudice and inequities present in labels. 

\begin{algorithm}[tb]
   \caption{Decoupled Confident Learning (DeCoLe)}\label{alg}
\begin{algorithmic}
   \STATE {\bfseries Input:} Noisy dataset $\boldsymbol{D} \coloneqq (\boldsymbol{x}, \tilde y) ^n$,  group indicator $g$, 
   \STATE initialize a set of classifiers $\{$C$_{1},...,$ C$_{k}\}$ 
   \FOR{$i=1$ {\bfseries to} $k$}
   \STATE {\bfseries Part 1: Estimating $p(\boldsymbol{x})$}
   \STATE $C_i$.fit$(\boldsymbol{x}_{g_i}, \tilde{y} )$ where $ \boldsymbol{x} \in g_i$ 
   \STATE $\hat p(\boldsymbol{x}_{gi}) \leftarrow $ $C_i$.predict\_crossval\_prob $(\tilde{y} = 1| \boldsymbol{x}_{g_i})$
   \STATE {\bfseries Part 2: Estimating the thresholds}
   \STATE LB$_{g_i} = $ LB$(y^*=1, g = g_i) = E_{\boldsymbol{x}\in \tilde y = 1, g = g_i}[\hat p(x)]$   
   \STATE UB$_{g_i} = $ UB$(y^*=0, g = g_i) = E_{\boldsymbol{x}\in \tilde y = 0, g = g_i} [\hat p(x)]$
   \STATE {\bfseries Part 3: Pruning}
   \STATE Remove $(\boldsymbol{x}_{g_i},\tilde{y}) \in \boldsymbol{D} $ where $\tilde y=1,  \hat p(\boldsymbol{x}_{g_i}) < $ UB$_{g_i}$
   \STATE Remove $(\boldsymbol{x}_{g_i},\tilde{y}) \in \boldsymbol{D} $ where $\tilde y=0,  \hat p(\boldsymbol{x}_{g_i}) > $ LB$_{g_i}$ 
   \ENDFOR
\end{algorithmic}
\end{algorithm}

\subsection{Decoupled Confident Learning (DeCoLe)} \label{DeCoLe}
We propose Decoupled Confident Learning (DeCoLe). At a high level, DeCoLe tackles the goal of pruning biased labels by training decoupled classifiers for each group $g_i, i\in \{0,.., k\}$, and applying a series of confident learning procedures in  parallel, in order to separately identify noise for each group. 
For each group, the algorithm identifies \emph{pruning thresholds}, which are a set of lower bounds (LB$_{g_i}$) and upper bounds (UB$_{g_i}$) of predicted probabilities.
The key challenge is to find out the regions where we can confidently determine that any instances within group $g_i$ with predicted probabilities above the LB$_{g_i}$ belong to the positive class, while any instances with predicted probabilities below the UB$_{g_i}$ belong to the negative class. Consequently, we prune instances that are confidently predicted to belong to the positive class, yet have an observed negative label, and vice versa. 
The main pruning procedure consists of three steps: 

\begin{enumerate}
    \item Train a separate predictive model $C_i$ for each group $g_i$, and obtain out-of-sample predicted probabilities $\hat p(\boldsymbol{x}_{g_i}) = \hat p(\tilde y = 1; \boldsymbol{x}_{g_i}, C_i)$, where $\boldsymbol{x}_{gi}$ denotes the instances $\boldsymbol{x}$ that belong to group $g_i$, and $\hat p(\tilde y = 1; \boldsymbol{x}, C_i)$ is the estimated probability of instance $\boldsymbol{x}$ belonging to class 1 according to the classifier $C_i$.
\item For each group $g_i$, estimate the upper bound (UB$_{g_i}$) threshold and lower bound (LB$_{g_i}$) threshold that can be used to identify instances that are inferred to have an erroneous observed label. Formally,  
\begin{align*}
    UB_{g_i} =  UB(y^*=0, g = g_i) = E_{\boldsymbol{x}\in \tilde y = 0, g = g_i} [\hat p(\boldsymbol{x})]\\
    LB_{g_i} =  LB(y^*=1, g = g_i) = E_{\boldsymbol{x}\in \tilde y = 1, g = g_i} [\hat p(\boldsymbol{x})]
\end{align*}
\item Prune instances whose observed label $\tilde y = 0$, and predicted probability $\hat p(\boldsymbol{x}_{g_i}) \geq $ LB$_{g_i}$. Analogously, prune instances whose observed label $\tilde y = 1$ and predicted probability $\hat p(\boldsymbol{x}_{g_i}) \leq $  UB$_{g_i}$. 
\end{enumerate}

The detailed algorithm can be found in \cref{alg}. When there is group and class-conditional noise, decoupling---training individual models for each group---disentangles the noise structure. For each predictive model for $g_i$, the noise becomes class conditional. At that stage, the theoretical guarantees of confident learning provided by~\citet{northcutt2021confident} are inherited by each classifier $C_i$. 

\section{Experiments}\label{experiments}
We first test the efficacy of Decoupled Confident Learning (DeCoLe) on a synthetic setting, and show that it improves pruning recall and precision across groups. This allows us to illustrate how the method works and validate its performance in a context where we have full control over the relationship between $y^*$ and $\tilde{y}$. We then apply DeCoLe in the context of hate speech detection, a domain where label bias has been recognized to be an important problem. We compare our approach with the classical Confident Learning (CL) algorithm ~\citep{northcutt2021confident}, and with random sampling (Random). In \cref{Simu_describ}, we motivate and describe in detail the synthetic dataset, and present results on this dataset in \cref{Simu_results}. In \cref{Hate_Speech_describ}, we present the details of the empirical dataset of hate speech detection, and demonstrate that DeCoLe effectively prunes erroneous instances and mitigates false negatives in hate speech labels, yielding better results than classical CL.

We assess the quality of pruning and the fairness metrics of a cleaned dataset post-pruning. To evaluate the quality of pruning, we focus on measuring the recall and precision of label error detection. Additionally, we examine the remaining label bias in the cleaned dataset by measuring false positive and false negative rates of the observed labels $\tilde y$ vs. the ground truth labels $y^*$. 



\begin{figure}[ht]
\vskip 0.1in
\begin{center}
\centerline{\includegraphics[width=0.95\columnwidth]{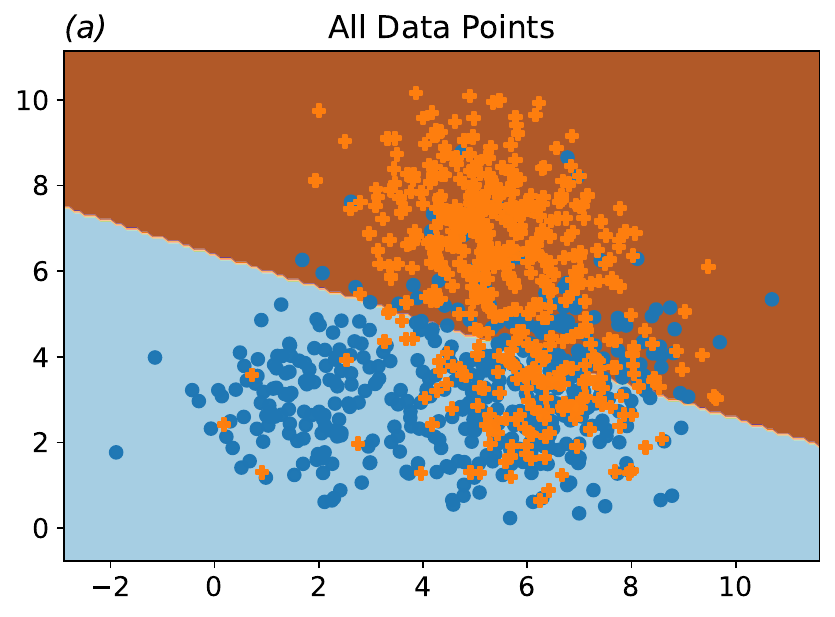}}
\centering
\begin{minipage}[b]{0.5\columnwidth}
\centering\includegraphics[scale = 0.31]{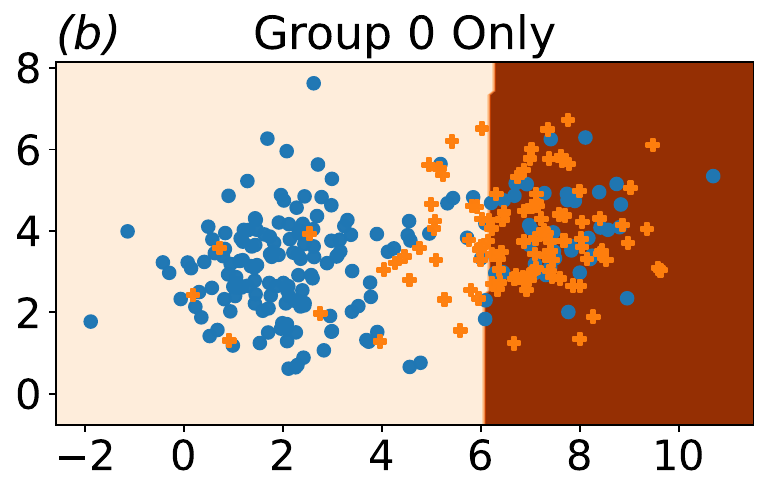} 
\end{minipage}%
\begin{minipage}[b]{0.5\columnwidth}
\centering\includegraphics[scale = 0.31]{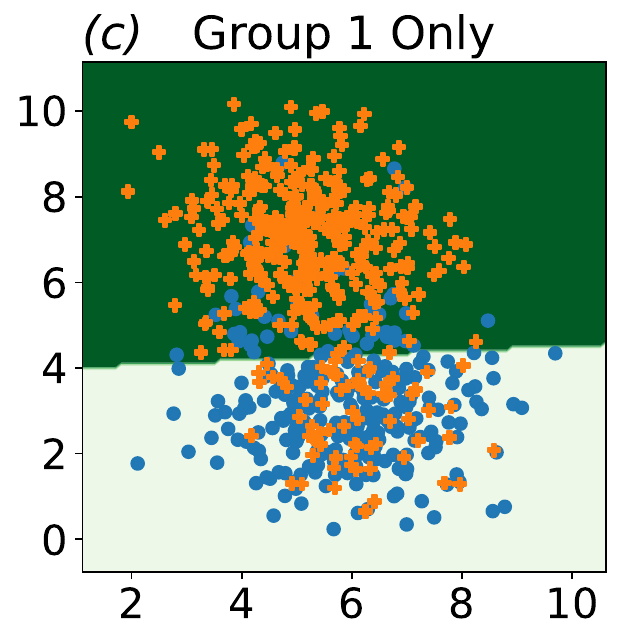}
\end{minipage}%
\caption{Dataset generated with the group and class-conditional noise. $y^*$ are represented by orange-thickened plus signs and negative instances are denoted by blue-filled circles. Figure (a) encompasses all data points, while (b) only includes instances belonging to $g_0$, and (c) only includes instances belonging to $g_1$. Group $g_1$, which constitutes 70\% of the instances, is the majority group. Observed labels for $g_0$ suffer from a high false negative rate, while those for $g_1$ have a high rate of false positives.}
\label{fig: simulation_distribution}
\end{center}
\vskip -0.2in
\end{figure}

\begin{figure*}[ht]
\begin{minipage}[c]{0.9\columnwidth}
\centering\includegraphics[scale = 0.48]{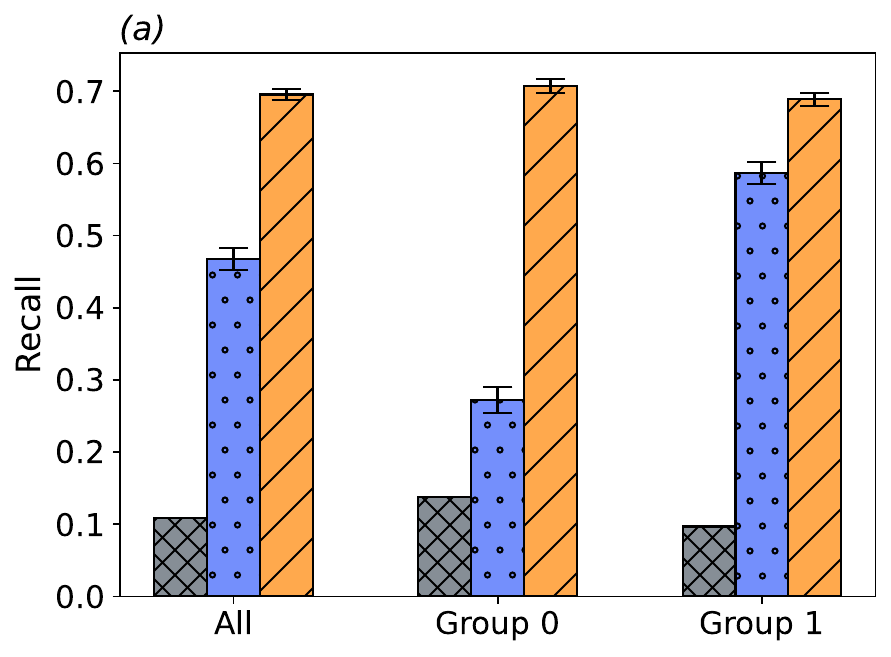} 
\end{minipage}%
\begin{minipage}[c]{1.1\columnwidth}
\centering\includegraphics[scale = 0.48]{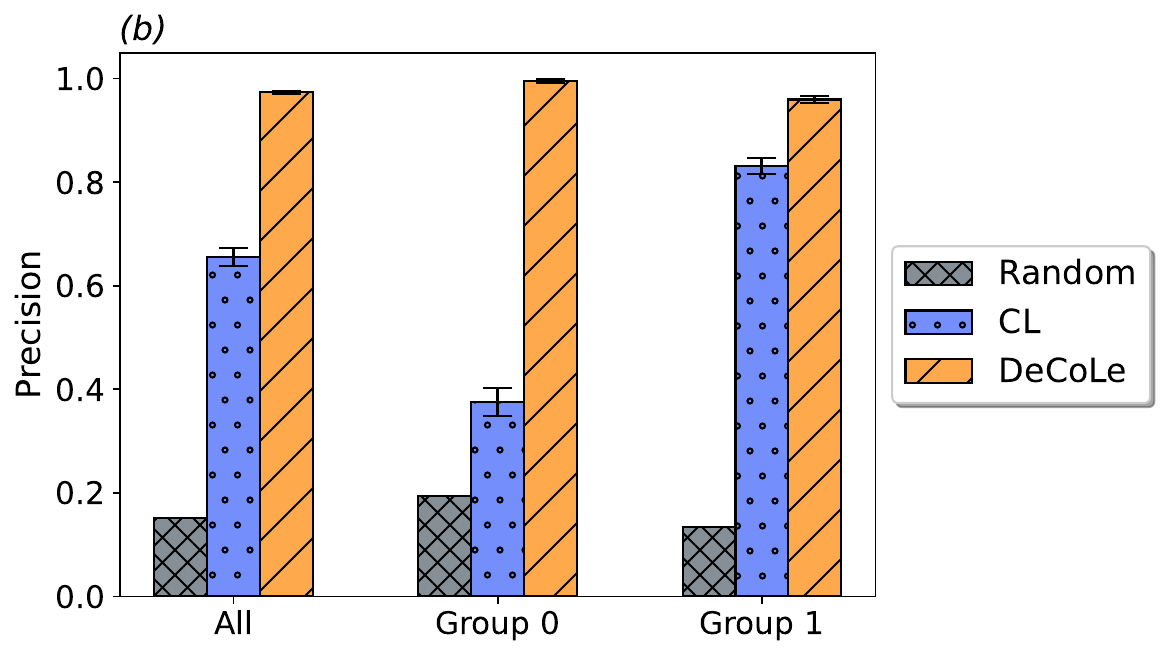}
\end{minipage}%
\caption{Pruning recall (a) and pruning precision (b) over all instances, group $g_0$ instances, group $g_1$ instances. Striped orange bars represent DeCoLe, while dotted blue bars and grid grey bars represent classical confident learning (CL) and random sampling, respectively. DeCoLe significantly outperforms CL in all scenarios, with particularly higher recall and precision for identifying erroneous labels of group $g_0$, the disadvantaged group.}
\label{fig:Simu8 Recall and Precision}
\end{figure*}

\subsection{Synthetic Experiments} \label{sec:synthetic}
In this section, we present a preliminary validation, showing that DeCoLe effectively identifies erroneous labels and mitigates label bias in a synthetic dataset. We illustrate how, compared to classical confident learning, DeCoLe significantly improves pruning recall, pruning precision, as well as fairness metrics of the cleaned dataset. 
\subsubsection{Data Generation}\label{Simu_describ}
We create a synthetic dataset with group and class-conditional noise rates, which allows us to  have full control of the relationship between observed labels $\tilde y$ and latent ground truth labels $y^*$. 

We introduce a synthetic population consisting of $N$ = 10000 instances, each associated with covariates $\boldsymbol{X} \in \mathbb{R}^2$, a binary group membership $g \in \{0, 1\}$, an outcome of interest $y^* \in \{0, 1\}$, and an observed label $\tilde y \in \{0, 1\}$. We consider group imbalance, a widely recognized issue in algorithmic fairness \cite{mitchell2018prediction}, by creating a predominant group ($g = 1$) representing 70\% of the total population. To account for differential subgroup validity \cite{hunter1979differential, de2022algorithmic}, which denotes differences in the relationship between covariates and target labels across groups, we draw instances for different group and class combinations from bi-dimensional normal distributions with different means. We use the same standard deviation for all normal distributions.  Further details about the sampling of $\boldsymbol{X}$ and labels $y^* \in \{0, 1\}$ can be found in \cref{Appendix: simulation}. 

We generate observed labels $\tilde y$ with group and class-conditional noise, i.e. different error types for different groups. Suppose the positive class represents opportunities or goods, such as job offers. We assume group $g_0$, the minority group, is more likely to be affected by false negative labels, and group $g_1$, the majority group, benefits from false positive labels. We set 
$\pi_{1\_{g_0}} = 0.4$ and $\pi_{0\_{g_1}} = 0.2$.  
Additionally, we assume some level of noise for the remaining instances, and set $\pi_{0\_{g_0}} = 0.05$ and $\pi_{1\_{g_1}} = 0.05$. 

This simulation therefore consists of four clusters depicted in \cref{fig: simulation_distribution} (b) and (c), where positive observed labels are represented by orange thickened plus signs and negative observed labels are denoted by blue filled circles. \cref{fig: simulation_distribution} (b) corresponds to group $g_0$ instances, while \cref{fig: simulation_distribution} (c) corresponds to group $g_1$ instances. Combining \cref{fig: simulation_distribution} (b) and (c) together, we get the full picture of the four clusters in \cref{fig: simulation_distribution} (a). Assuming we were to use a linear classifier to differentiate between the two classes, it is not hard to find that linear classifiers for group $g_0$ only (\cref{fig: simulation_distribution} (b)) and for group $g_1$ only (\cref{fig: simulation_distribution} (c)) exhibit fundamental dissimilarity, being nearly orthogonal.  Additionally, when we fit one linear classifier for both groups, as depicted in \cref{fig: simulation_distribution} (a), it demonstrates differential subgroup validity, wherein its predictive accuracy is notably higher for the majority group compared to the minority group. Furthermore, the linear classifier in \cref{fig: simulation_distribution} (a) tends to misclassify positive group $g_0$ instances as negative and negative group $g_1$ instances as positive, reflecting how the predictive model may lean and amplify bias in the data labels. In the same way, this affects models used for pruning, and, as we show, results in poor pruning performance when we do not consider the group-specific nature of predictive relationships and label errors.

\subsubsection{Results and Analysis}\label{Simu_results}

We apply DeCoLe framework described in~\cref{alg} on the dataset generated in~\cref{Simu_describ}. We also consider the CL alternative as a baseline and include the performance of random pruning as reference. We use logistic regression as a base model, and generate 95\% confidence bounds via 5 runs on different seeds.

\begin{figure*}
\centering\includegraphics[scale = 0.48]{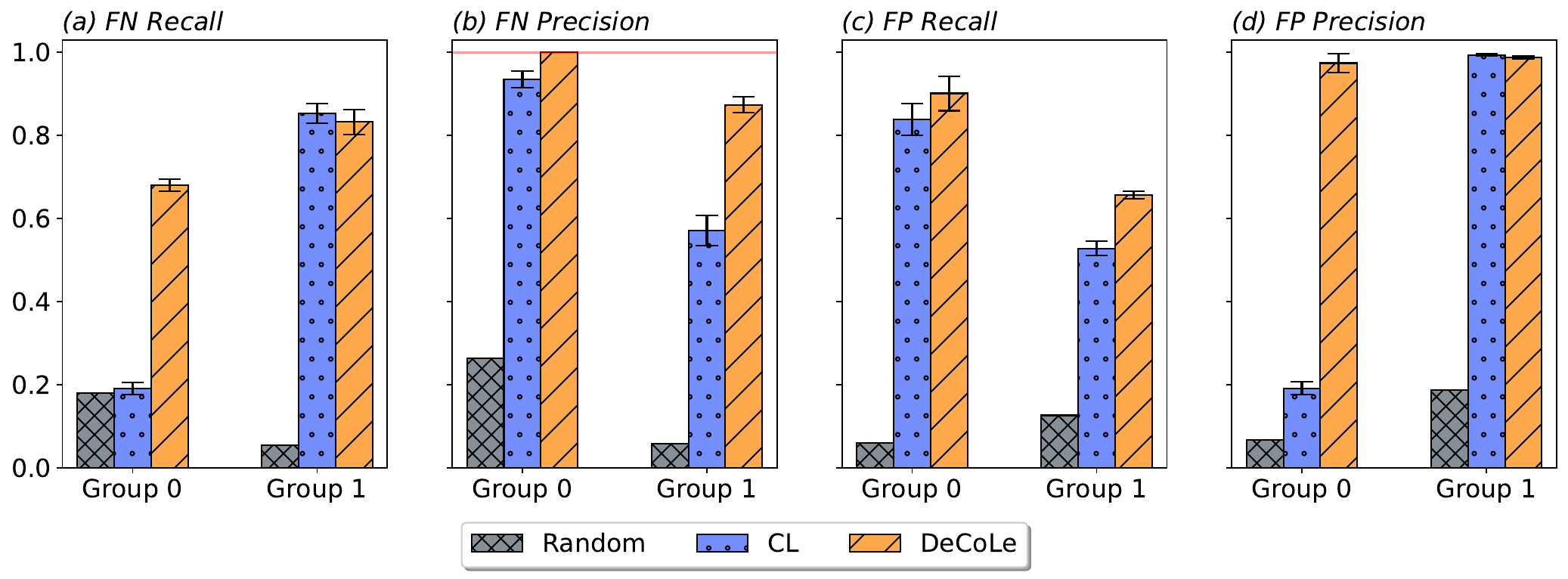} 
\caption{Pruning recall and precision of different error types over group $g_0$ instances and group $g_1$ instances: (a) false negatives recall; (b) false negatives precision; (c) false positives recall; (d) false positives precision. Striped orange bars represent DeCoLe, while dotted blue bars and grid grey bars represent classical confident learning (CL) and random sampling, respectively. According to (a) and (b), DeCoLe significantly improves recall and precision on identifying false negatives for group $g_0$. According to (c), DeCoLe significantly improves recall rate on identifying false positives for group $g_1$. The improvement on FN recall and FP precision for group $g_0$ is especially large. }
\label{fig:Simu8_FP_FN_Recall_and_Precision}
\end{figure*}

\begin{figure}[ht]
\vskip 0.1in
\begin{center}
\centerline{\includegraphics[width=0.95\columnwidth]{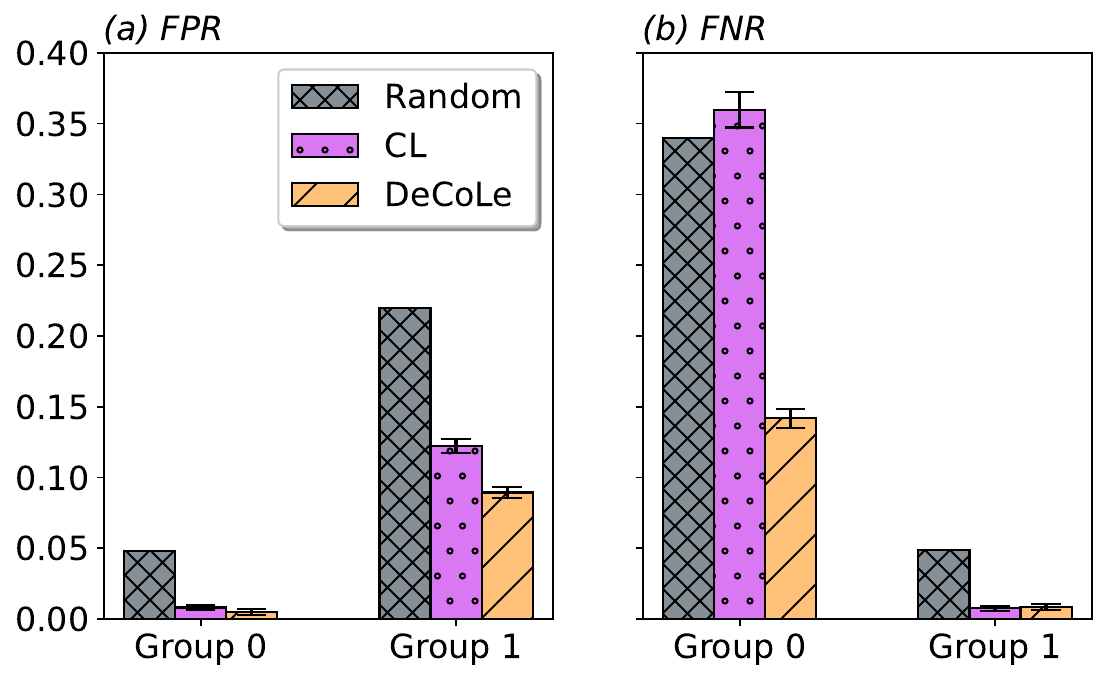}}
\centering
\caption{(a) False Positive Rates (FPR) and (b) False Negative Rates (FNR) over group 0 and group 1 instances of data after pruning employing random sampling (grid grey bars), CL algorithm (dotted orchid vars), and DeCoLe (striped pastel orange bars). DeCoLe substantially more capably mitigates both error types (false positives for group 1 and false negatives for group 0).}
\label{fig: Simu8_fair_metrics}
\end{center}
\vskip -0.2in
\end{figure}


\cref{fig:Simu8 Recall and Precision} (a) and (b) illustrate the pruning recall and precision, respectively. For each, we can assess these metrics overall, for group $g_0$, and for group $g_1$. Striped orange bars represent DeCoLe framework, while dotted blue bars and grid grey bars represent classical confident learning (CL) and random sampling, respectively. \cref{fig:Simu8 Recall and Precision} clearly demonstrates that \emph{DeCoLe significantly outperforms CL in all scenarios, with particularly remarkable higher accuracy in correctly identifying erroneous labels of group $g_0$, the disadvantaged group.}   

For a more nuanced view, \cref{fig:Simu8_FP_FN_Recall_and_Precision} shows the pruning precision and recall for different error types over the two groups. As we described in \cref{Simu_describ}, group $g_0$ mainly suffers from false negatives, while the main error for group $g_1$ labels is false positives. According to \cref{fig:Simu8_FP_FN_Recall_and_Precision} (a) and (b), DeCoLe significantly improves recall and precision on identifying  false negatives (FN) for group $g_0$. Additionally, according to \cref{fig:Simu8_FP_FN_Recall_and_Precision} (c), DeCoLe significantly improves recall rate on identifying false positives (FP) for group $g_1$. The improvement on FN recall and precision for group $g_0$ is especially large, which is important from a fairness perspective. Furthermore, according to \cref{fig:Simu8_FP_FN_Recall_and_Precision} (d), CL produces extraordinarily low FP precision for group $g_0$. There are very few (only 5\%) false positives in group $g_0$; such a low precision means that the CL algorithm pruned many positive instances in group $g_0$ even though those instances are true positives, which would further exacerbate the existing bias in the data.  

The central goal of pruning is to effectively detect inaccurate labels in order to yield a much cleaner dataset, preventing bias propagation through ML models trained on data with different error types for different groups. \cref{fig: Simu8_fair_metrics} depicts the quality of the label $\tilde y$ with respect to the label $y^*$ in the cleaned data. \cref{fig: Simu8_fair_metrics} (a) shows the FPR, the proportion of actual negatives ($y^*$) incorrectly labeled as positives ($\tilde y$);  \cref{fig: Simu8_fair_metrics} (b) shows the FNR, the proportion of actual positives ($y^*$) incorrectly labeled as negatives ($\tilde y$). The results show the errors in the data post-pruning, when pruning is done employing random sampling (grid grey bars), CL algorithm (dotted orchid bars), and DeCoLe (striped pastel orange bars). As in the other cases, the results are shown for both group $g_0$ and group $g_1$ instances. As shown in \cref{fig: Simu8_fair_metrics} (b), CL yields worse-than-random performance for the disadvantaged group, $g_0$. Across both plots, it is evident that \emph{DeCoLe is substantially more capable of mitigating the two most prominent error types (false positives for group $g_1$ and false negatives for group $g_0$) and thus is more suitable for preventing systematic bias present in labels.}

\begin{figure*}
\begin{minipage}[c]{0.9\columnwidth}
\centering\includegraphics[scale = 0.48]{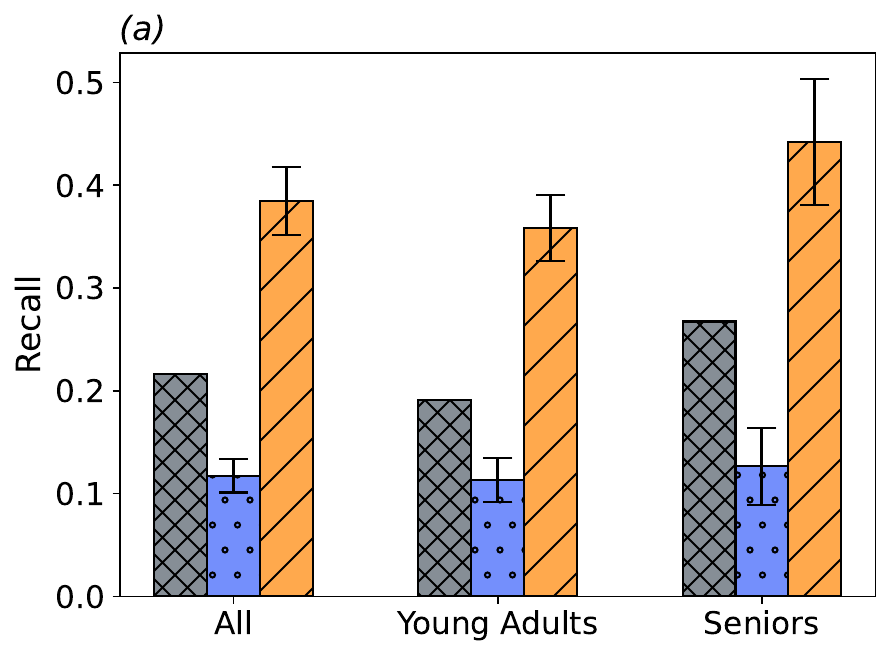} 
\end{minipage}%
\begin{minipage}[c]{1.1\columnwidth}
\centering\includegraphics[scale = 0.48]{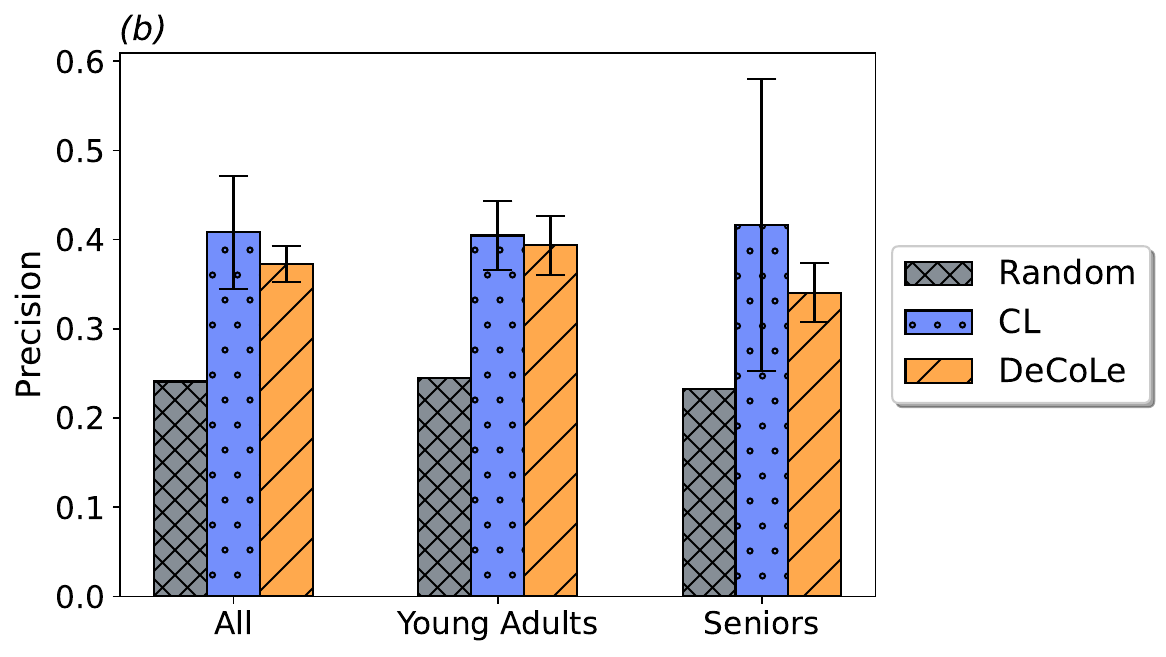}
\end{minipage}%
\caption{In the context of hate speech detection, pruning (a) recall rates and (b) precision rates over all instances, posts targeting young adults, and posts targeting seniors. Striped orange bars represent DeCoLe, while dotted blue bars and grid grey bars represent classical confident learning (CL) and random sampling, respectively. DeCoLe significantly outperforms CL in recall rates while yielding comparable precision rates.
}
\label{fig:Recall and Precision}
\end{figure*}


\subsection{Hate Speech Label Bias Mitigation}
\label{Hate_Speech_describ}

Hate speech causes significant harm. It is used to radicalize and recruit within extremist groups, incite violence, and even genocide~\cite{kennedy2020constructing}. However, labeling hate speech is challenging given that judgments of offensiveness depend on societal circumstances~\cite{sap2019risk}. Extremist groups also intentionally make their hate speech obscure to evade detection~\cite{kennedy2020constructing}. Crowdsourced annotations used for training automatic hate speech detection systems are prone to bias from varied annotator knowledge and perspectives in what constitutes hate speech~\cite{davani2022dealing}. What one person labels as hateful, another may see as benign, yielding conflicting labels for the same data~\cite{kennedy2020constructing, davani2022dealing}. Although majority votes are commonly used to aggregate multiple opinions, normative stereotypes embedded in society and homogeneity of the annotators' bias can easily lead to systematic labeling errors~\cite{davani2023hate, davani2022dealing}.

Recently, \citet{kennedy2020constructing} proposed a novel method based on Rasch Measurement Theory (RMT) to construct a less biased hate speech measure. Their measure articulates hate speech theoretically across eight dimensions (incite violence, humiliate, etc.), capturing the complexity of hate speech and limiting bias from oversimplification. Furthermore, by evaluating inter-rater reliability, they are able to remove inconsistent raters, correcting human judgment biases and promoting reliability. 
The researchers assessed the validity and reliability of their proposed measurement approach and found that it demonstrates high internal consistency, test-retest reliability, and construct validity. In summary, \citet{kennedy2020constructing} limits bias in labels for hate speech, but it involves a significantly more costly labeling process, as it requires labels for each instance across eight different dimensions, rather than one. The dataset also has the advantage of collecting the more common labels used for hate speech (directly asking if a post constitutes hate speech), and includes demographic information about the target group of the posts. Thus, the data contains a label $\tilde y$ and an improved label $y^*$ that we leverage to assess the performance of DeCoLe in a real-world dataset from an impactful domain. 


\subsubsection{Empirical Results} \label{Hate_Speech_results} 
To empirically validate that DeCoLe effectively identifies erroneous labels and mitigates label bias in data used to train hate speech detection systems, we utilized the newly generated RMT-based hate speech measure as our gold standard labels, $y^*$. The observed labels $\tilde y$, which may contain biases, were obtained from a single hate speech survey item~\cite{kennedy2020constructing, sachdeva2022measuring}. We conducted this validation specifically in the context of hate speech targeting two groups: young adults and seniors. We use random forest as the base model, and to ensure robustness, we performed five runs with different random seeds to obtain a 95\% confidence bound. Our empirical findings demonstrate that DeCoLe outperforms CL algorithm by significantly improving the recall rate of erroneous labels and more effectively mitigating false negatives for both groups.

\begin{figure}[ht]
\vskip 0.1in
\begin{center}
\centerline{\includegraphics[width=0.95\columnwidth]{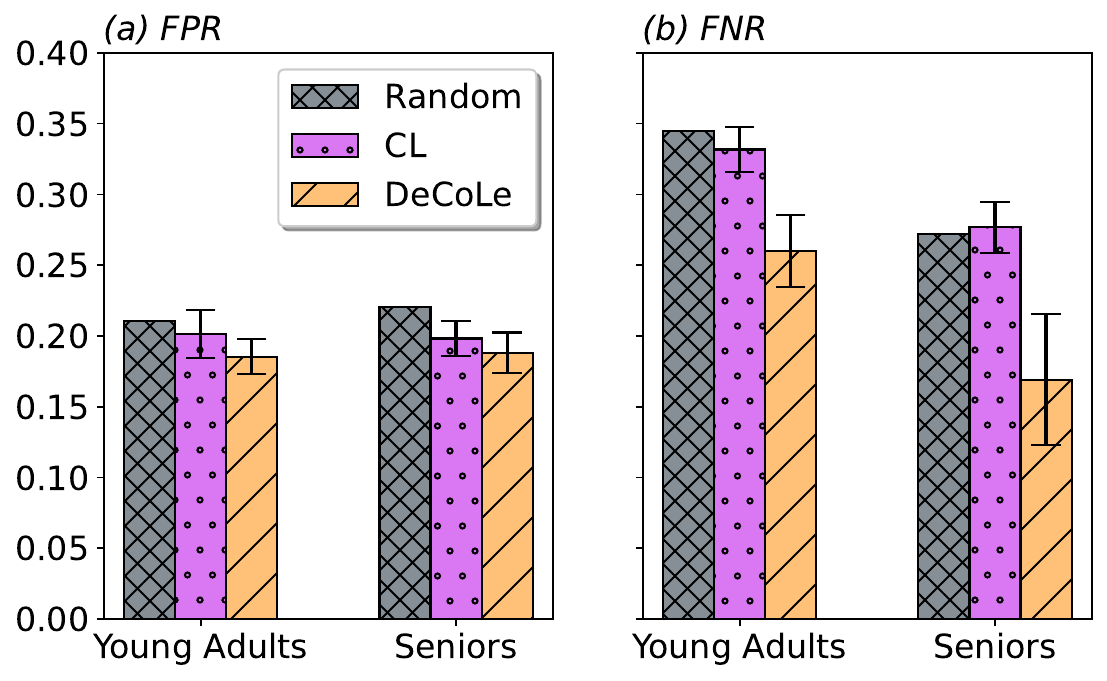}}
\centering
\caption{(a) False positive rates and (b) false negative rates of the dataset after pruning using random sampling (grid grey bars), CL algorithm (dotted orchid bars), and DeCoLe (striped pastel orange bars) framework. The grid grey bars could be understood as representing the error rates of the dataset before pruning. According to (b), DeCoLe significantly reduces false negatives for both posts targeting young adults and posts targeting Seniors, surpassing the performance of CL algorithm.}
\label{fig: fair_metrics}
\end{center}
\vskip -0.2in
\end{figure}


\cref{fig:Recall and Precision} (a) and (b) show the pruning recall (a) and pruning precision (b) for all posts, posts targeting young adults, and posts targeting seniors. The striped orange bars represent the results from DeCoLe, while the dotted blue bars and grid grey bars represent CL approach and random sampling, respectively. It can be observed that while both CL and DeCoLe yield comparable precision, DeCoLe demonstrates significant improvement in recall compared to both CL and random sampling.

Furthermore, \cref{fig: fair_metrics} contains information about the quality and fairness of the cleaned dataset after pruning with different methods. \cref{fig: fair_metrics}  (a) displays the false positive rates and \cref{fig: fair_metrics} (b) the false negative rates of observed labels in the cleaned dataset after pruning using random sampling (grid grey bars), CL (dotted orchid bars), and DeCoLe algorithm (striped pastel orange bars). The results are disaggregated for posts targeting young adults, and posts targeting seniors. Note that random sampling maintains the same rate of errors, thus, the grid grey bars can also be understood as representing the error rates of the dataset before pruning. From \cref{fig: fair_metrics} (b), it can be observed that DeCoLe significantly reduces false negatives for both posts targeting young adults and posts targeting seniors, surpassing the performance of CL algorithm.

\section{Conclusion} \label{conclusion}
While there is a growing awareness of the presence of label bias in supervised learning systems, particularly those used to guide high-stake decisions, methods that are specifically designed for mitigating label bias remain insufficient. To address this pressing issue, we propose a novel approach called Decoupled Confident Learning (DeCoLe), a pruning method that mitigates label bias. Specifically, DeCoLe improves upon existing noise-mitigation alternatives by accounting for the fact that noise may be group- and class-conditioned. 
This type of label bias arises when the likelihood that a label is incorrect is influenced by both the group membership and the ground truth class. For instance, in the context of hate speech, it has been shown that labelers' assessment of hate speech depends on the stereotypes they have about a given group~\citep{davani2023hate}. Our experimental results, which focus on the hate speech domain, 
validate the effectiveness of DeCoLe in pruning erroneous instances and mitigating group-specific false negatives associated with hate speech labels. 

Future research endeavors should focus on the development of methodologies capable of handling other forms of label bias structures. While DeCoLe has primarily focused on mitigating group and class-conditional noise, there is a need for novel methodologies that can address label noise patterns conditioned on other covariates. 
By extending existing approaches to encompass diverse sources of label noise, researchers can advance the field's understanding and ability to mitigate biases arising from a wider range of factors.

\section{Acknowledgements}
This research was supported by the Machine Learning Laboratory\footnote{https://ml.utexas.edu/}, and Good Systems\footnote{http://goodsystems.utexas.edu/}, a UT Austin Grand Challenge to develop responsible AI technologies.

\bibliography{DeCoLe}
\bibliographystyle{icml2023}

\newpage
\appendix
\onecolumn
\section{Appendix: Synthetic Data Generation}\label{Appendix: simulation}
We describe the details of the simulated 4 clusters for different class and group combinations here. 

We introduce a bi-dimensional ($\boldsymbol{X} \in \mathbb{R}^2$) synthetic population ($N$ = 10000) divided into two groups ($g \in \{0, 1\}$), and assume group 1 is the majority group that account for 70\% of the population. Specifically, for group 0, we sample the attributes of class 0 instances $\boldsymbol{X}_{g=0, y^* =1} \in \mathbb{R}^2$ from normal distribution $\mathcal{N}((\mu_{x_1} = 2, \mu_{x_2} = 3),\sigma ^{2})$, and sample the attributes of class 1 instance $\boldsymbol{X}_{g=0, y^* =0} \in \mathbb{R}^2$ from normal distribution $\mathcal{N}((\mu_{x_1} = 7, \mu_{x_2} = 4),\sigma^{2})$. Similarly, we sample the attributes of group 1 class 0 instance $\boldsymbol{X}_{g=1, y =0} \in \mathbb{R}^2$ from $\mathcal{N}((\mu_{x_1} = 6, \mu_{x_2} = 3),\sigma ^{2})$, and sample group 1 class 1 instance $\boldsymbol{X}_{g=1, y =1} \in \mathbb{R}^2$ from normal distribution $\mathcal{N}((\mu_{x_1} = 5, \mu_{x_2} = 7),\sigma ^{2})$. We set $\sigma = 1.2$ for all distributions.

\end{document}